\documentclass[letterpaper]{article} % DO NOT CHANGE THIS
\usepackage{aaai2026}  % DO NOT CHANGE THIS
\usepackage{times}  % DO NOT CHANGE THIS
\usepackage{helvet}  % DO NOT CHANGE THIS
\usepackage{courier}  % DO NOT CHANGE THIS
\usepackage[hyphens]{url}  % DO NOT CHANGE THIS
\usepackage{graphicx} % DO NOT CHANGE THIS
\usepackage{natbib}  % DO NOT CHANGE THIS AND DO NOT ADD ANY OPTIONS TO IT
\usepackage{caption} % DO NOT CHANGE THIS AND DO NOT ADD ANY OPTIONS TO IT
\usepackage{algorithm}
\usepackage{algorithmic}

\usepackage[colorinlistoftodos]{todonotes} % this one to show todos
\usepackage{newfloat}
\usepackage{listings}
\DeclareCaptionStyle{ruled}{labelfont=normalfont,labelsep=colon,strut=off} % DO NOT CHANGE THIS
\floatstyle{ruled}
\newfloat{listing}{tb}{lst}{}
\floatname{listing}{Listing}
\title{Principles2Plan: LLM-Guided System for \\Operationalising Ethical Principles into Plans}
\author {
    Tammy Zhong,
    Yang Song,
    Maurice Pagnucco
}
\affiliations {
    School of Computer Science and Engineering, University of New South Wales,
Sydney, Australia\\
    \{tammy.zhong,yang.song1,morri\}@unsw.edu.au
}

\begin{document}

\maketitle

\begin{abstract}

Ethical awareness is critical for robots operating in human environments, yet existing automated planning tools provide little support. Manually specifying ethical rules is labour-intensive and highly context-specific. We present \emph{Principles2Plan}, an interactive research prototype demonstrating how a human and a Large Language Model (LLM) can collaborate to produce context-sensitive ethical rules and guide automated planning. A domain expert provides the planning domain, problem details, and relevant high-level principles such as beneficence and privacy. The system generates operationalisable ethical rules consistent with these principles, which the user can review, prioritise, and supply to a planner to produce ethically-informed plans. To our knowledge, no prior system supports users in generating principle-grounded rules for classical planning contexts. Principles2Plan showcases the potential of human-LLM collaboration for making ethical automated planning more practical and feasible.

\end{abstract}

\section{Introduction}

The deployment of robots around people raises the challenge of ensuring that their actions achieve goals while respecting ethical principles. High-level ethical principles, such as beneficence, depend heavily on context. For example, in an autonomous vehicle scenario, a passenger needing urgent medical attention may justify taking an unauthorised shortcut to reach the hospital quickly, whereas for a leisure trip, following standard traffic rules may be preferable to avoid unnecessary risk. In both cases, the principle applies, yet the resulting actions differ. This illustrates a key challenge: interpreting abstract ethical principles in real-world scenarios is nuanced, context-dependent, and often controversial, making fully automated ethical planning difficult. 
We aim to develop an interactive software platform, based on existing work, that encourages human-machine collaboration to interpret these
principles in a given classical planning problem and generate plans that not only achieve goals, but also consider the ethics of the plan that achieves such goals.

\emph{Computational Machine Ethics (CME)} approaches are often divided into top-down, bottom-up, and hybrid approaches. Top-down methods \cite{vanderelst_architecture_2018,pagnucco_epistemic_2021,grandi_logic-based_2023} specify rules or guidelines in advance, ensuring transparency but lacking adaptability. Bottom-up approaches \cite{jiang_delphi_2021,li_reinforcement_2025} rely on data to infer ethical behaviour, trading off interpretability for flexibility. Hybrid approaches \cite{allen_artificial_2005,ramanayake_implementing_2024} attempt to combine these strengths, but typically still require extensive manual effort to encode ethical rules or examples. Advances in large language models (LLMs) offer a practical means to reduce the manual effort of encoding such rules or examples, which we consider in a planning context.

Recent work has explored incorporating LLMs into automated planning in various ways \cite{pallagani_prospects_2024}. Beyond attempts to use LLMs to generate plans directly, they have been applied to facilitate planning processes, including model construction \cite{oswald_large_2024}, human--LLM collaboration \cite{wu_integrating_2023}, and translation of natural language into structured languages \cite{ahn_as_2022,liu_llmp_2023,favier_collaborative_2025,zhong_computational_2025}.
Few contemporary studies leverage LLMs to support automated planning with explicit specifications \cite{favier_collaborative_2025,zhong_computational_2025}. \citet{favier_collaborative_2025} use LLMs to decompose and encode general natural language constraints in PDDL3, while \citet{zhong_generation_2025} translate high-level ethical principles into context-specific rules represented as action costs in PDDL. Although the latter targets ethics---an underexplored area in automated planning---it lacks a user-facing interface, which \citet{favier_collaborative_2025} provides.
We present \emph{Principles2Plan}, a prototype that enables users to generate ethical plans. While prior work lies at the intersection of users, LLMs, and automated planning, no existing system supports collaborative human--LLM refinement and operationalisation of ethical principles. Principles2Plan addresses this gap by integrating an interactive interface with the pipeline introduced by \citet{zhong_generation_2025}.

Principles2Plan is a prototype that leverages LLMs and human oversight to incorporate ethical considerations into automated planning. Building on the human-in-the-loop pipeline introduced in \cite{zhong_generation_2025}, the system takes user input, which an LLM uses to generate context-specific ethical rules from high-level principles. Users can then refine and prioritise these rules before supplying them to a PDDL classical planner.
This design makes explicit how abstract principles are operationalised into actionable rules to guide planning, enabling transparent and ethically informed plans in real time. By emphasising interactivity and usability, Principles2Plan contributes a practical system demonstration of how LLMs can bridge the gap between high-level principles and lower-level automated planning, showcasing a novel research direction in CME.

\section{System Overview}

\begin{figure}[t]
    \centering
    \includegraphics[width=0.9\columnwidth]{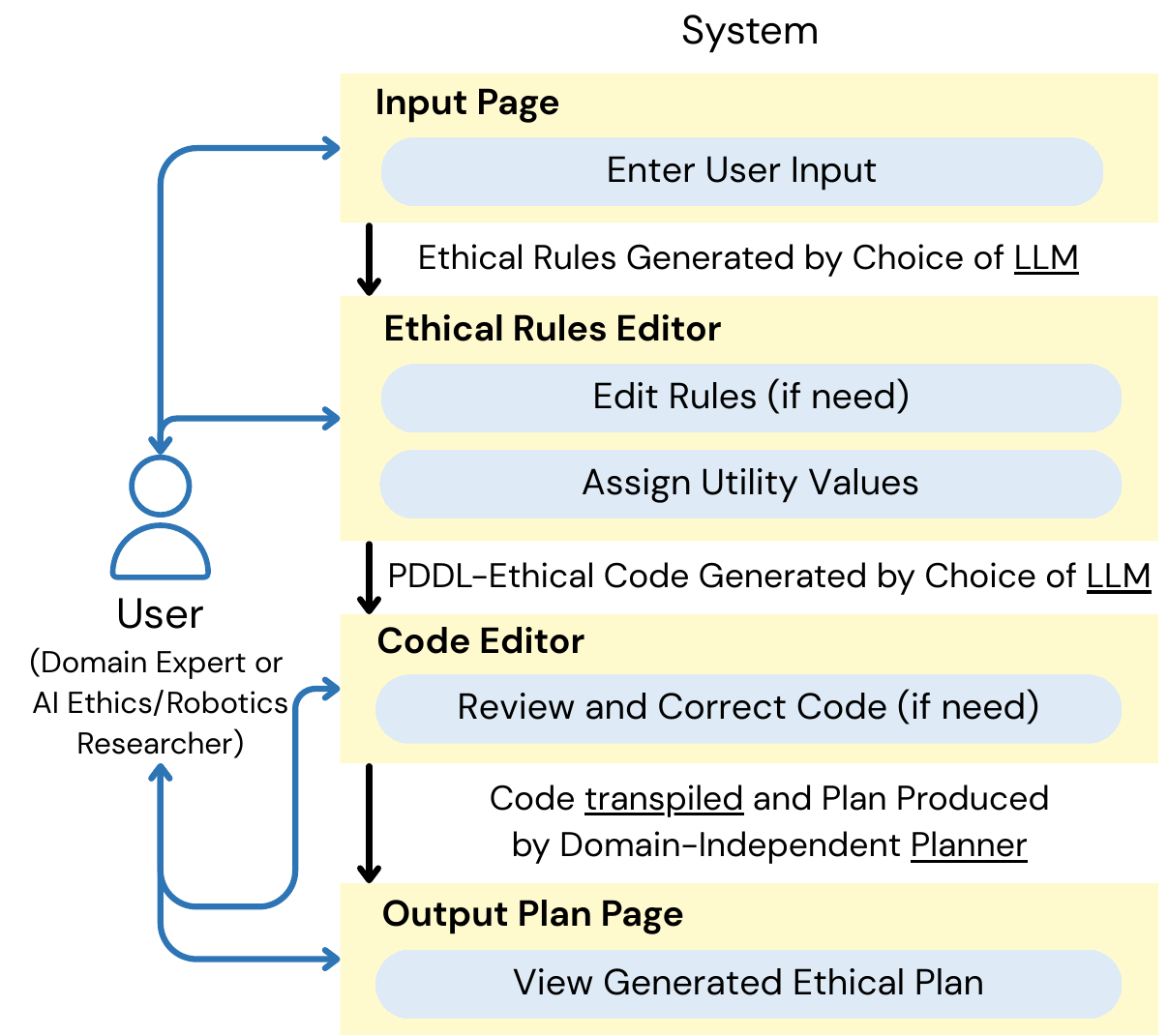} 
    \caption{Overall user/system flow.}
    \label{fig:flow}
\end{figure}

To generate an ethical plan from a planning problem and high-level ethical principles, Principles2Plan guides users through four steps on dedicated pages: providing input, reviewing and prioritising generated rules, and reviewing code before producing an ethically-informed plan. Figure~\ref{fig:flow} illustrates this process from the user's perspective, which we describe in detail in this section. 
The intended users of the system are domain experts in ethically-sensitive domains, AI ethics and robotics researchers, and anyone interested in the intersection of ethics, LLMs, and automated planning.
As the process includes reviewing code and generating plans, users are assumed to have a basic understanding of automated planning and familiarity with PDDL\footnote{ https://planning.wiki/guide/whatis/pddl} (a standardised language used in planning) and PDDL-Ethical (an extension for ethical constructs) \cite{jedwabny_preference-based_2022}. 
We recognise that intended users are unlikely to have technical knowledge of planning and PDDL; minimising the need for such expertise remains a challenge for future work.

\textbf{Input Page}
The input page of Principles2Plan lets users start generating ethically-informed plans by providing key problem information. These inputs drive the system to generate context-specific ethical rules in natural language, following a structure defined in \cite{zhong_generation_2025}. Each rule includes \emph{ethical features}, representing positive or negative ethical characteristics of the rule (e.g., dishonesty as a negative feature). Users can upload and preview their \textit{problem.pddl} and \textit{domain.pddl} files. The user also specifies the initial state, assumptions about the problem or domain, and high-level ethical principles to guide rule generation. Finally, the user can select a preferred model. The system then processes all inputs and prompts the LLM to produce relevant ethical rules in real time.

To help users explore and experiment with the system, Principles2Plan provides multiple example problems across three ethically-sensitive domains: autonomous vehicles, elderly care, and firefighting/rescue. Users can select these examples to populate the input fields directly. 

\textbf{Ethical Rules Editor}
Since ethical rules generated by an LLM may be inconsistent or imperfect, the next step allows users to review and refine them. Users can add missing rules, remove inappropriate ones, and modify existing rules. To support this process, the system provides explanations from the LLM, detailing the reasoning behind why each rule was generated based on the problem and specified ethical principle(s).
Once users are satisfied with the rules, they can prioritise them by assigning a significance level (1–5) to each ethical feature associated with a rule. The system highlights positive and negative features, allowing users to click and adjust their importance easily. These rules are then fed into the LLM to generate PDDL-Ethical code, which users review on the following page.

\textbf{Code Editor}
On the code editing page, users review the syntax-highlighted PDDL-Ethical code generated from the natural language ethical rules. The code is then transpiled (using the method from \cite{jedwabny_preference-based_2022}) into raw PDDL with action costs and submitted to a domain-independent classical planner (Fast Downward). A view of the ethical rules from the previous page is provided alongside to support cross-checking, helping users ensure correctness and consistency between the rules and the code.

\textbf{Output Plan Page}
The plan generated with ethical rules and another produced by the same planner using the original problem and domain files are displayed side-by-side, allowing users to directly evaluate the impact of the ethical rules.

One may question the practicality and performance of LLM-generated outputs here and whether they add more work for the user. 
The performance of the method has been evaluated with DeepSeek-R1-Distill-Llama-70B in \cite{zhong_computational_2025} using metrics including Sentence-BERT similarity (0.82) for generated rules and code generation success rate (82.2\%). While these results are not exceptional, they indicate a promising direction.
As this is the first implemented prototype of its kind, it may require more human intervention in its current form. We are optimistic that future iterations will improve the balance of collaboration between humans and LLMs.

\section{Conclusion}

Principles2Plan is a novel prototype that enables ethically-aware automated planning by combining human guidance with LLMs. Users can generate, refine, and prioritise context-specific ethical rules to produce transparent and ethically-informed plans in real time. Future work will enhance human-LLM collaboration through iterative dialogue and suggestions. Overall, Principles2Plan serves as a hands-on platform for generating ethical plans and for researchers to experiment with interactive ethical decision-making.

\bibliography{aaai2026}

\end{document}